\pdfoutput=1
\documentclass[11pt]{article}

\usepackage[]{acl}
\usepackage{graphicx}
\usepackage{lipsum}
\usepackage{times}
\usepackage{amsmath}
\usepackage{latexsym}

\usepackage[T1]{fontenc}
\usepackage[utf8]{inputenc}

\usepackage{microtype}

\title{Findings of the Shared Task on Offensive Span Identification from Code-Mixed Tamil-English Comments }

\author{Manikandan Ravikiran\textsuperscript{$\dagger$}\thanks{\ \ Corresponding Author}\ , Bharathi Raja Chakravarthi\textsuperscript{$\ddagger$}, Anand Kumar Madasamy\textsuperscript{$\star$} \\
{\bf Sangeetha Sivanesan \textsuperscript{$\circ$}, Ratnavel Rajalakshmi \textsuperscript{$\oplus$}, Sajeetha Thavareesan\textsuperscript{$\diamond$}} \\
{\bf Rahul Ponnusamy \textsuperscript{$\ominus$}, Shankar Mahadevan\textsuperscript{$\bowtie$}
}\\ 
 \textsuperscript{$\dagger$}Georgia Institute of Technology, Atlanta, Georgia\\
\textsuperscript{$\ddagger$}Data Science Institute, National University of Ireland Galway\\
 \textsuperscript{$\star$}National Institute of Technology Karnataka Surathkal, India\\
\textsuperscript{$\circ$}National Institute of Technology, Trichy, India \\
\textsuperscript{$\oplus$}Vellore Institute of Technology, Chennai, India \\
\textsuperscript{$\diamond$}Eastern University, Sri Lanka \\
\textsuperscript{$\ominus$}Indian Institute of Information Technology and Management, Kerala, India \\
\textsuperscript{$\bowtie$}Thiagarajar College of Engineering, Madurai, India \\
\texttt{mravikiran3@gatech.edu, bharathi.raja@insight-centre.org} 
\\}

\begin{document}
\maketitle
\vspace{2cm}
\begin{abstract}
Offensive content moderation is vital in social media platforms to support healthy online discussions. However, their prevalence in code-mixed Dravidian languages is limited to classifying whole comments without identifying part of it contributing to offensiveness. Such limitation is primarily due to the lack of annotated data for offensive spans. Accordingly, in this shared task, we provide Tamil-English code-mixed social comments with offensive spans. This paper outlines the dataset so released, methods, and results of the submitted systems.
\end{abstract}

\section{Introduction}

Combating offensive content is crucial for different entities involved in content moderation, which includes social media companies as well as individuals \cite{kumaresan2021findings,chakravarthi-muralidaran-2021-findings}. To this end, moderation is often restrictive with either usage of human content moderators, who are expected to read through the content and flag the offensive mentions \cite{arsht_etcovitch_2018}. Alternatively, there are semi-automated and automated tools that employ trivial algorithms and block lists \cite{Jhaver2018OnlineHA}. Though content moderation looks like a one-way street, where either it should be allowed or removed, such decision-making is fairly hard. This is more significant, especially on social media platforms, where the sheer volume of content is overwhelming for human moderators especially. With ever increasing offensive social media contents focusing "racism", "sexism", "hate speech", "aggressiveness" etc. semi-automated and fully automated content moderation is favored \cite{priyadharshini2021overview,chakravarthi2020overview,emotion-acl}. However, most of the existing works ~\cite{Zampieri2020SemEval2020T1,lti-acl,speech-acl,comdet-acl} are restricted to English only, with few of them permeating into research that focuses on a more granular understanding of offensiveness.

Tamil is a agglutinative language from the Dravidian language family dating back to the 580 BCE \cite{sivanantham2019keeladi}. It is widely spoken in the southern state of Tamil Nadu in India, Sri Lanka, Malaysia, and Singapore.  Tamil is an official language of Tamil Nadu, Sri Lanka, Singapore, and the Union Territory of Puducherry in India. Significant minority speak Tamil in the four other South Indian states of Kerala, Karnataka, Andhra Pradesh, and Telangana, as well as the Union Territory of the Andaman and Nicobar Islands \cite{9606025,8300346,7946522,9063341,9185369,9342640,9605839}. It is also spoken by the Tamil diaspora, which may be found in Malaysia, Myanmar, South Africa, the United Kingdom, the United States, Canada, Australia, and Mauritius. Tamil is also the native language of Sri Lankan Moors. Tamil, one of the 22 scheduled languages in the Indian Constitution, was the first to be designated as a classical language of India \cite{Subalalitha2019,srinivasan2019automated,narasimhan2018porul}. Tamil is one of the world's longest-surviving classical languages. The earliest epigraphic documents discovered on rock edicts and "hero stones" date from the 6th century BC. Tamil has the oldest ancient non-Sanskritic Indian literature of any Indian language \cite{anita2019building,anita2019approach,subalalitha2018automatic}.  Despite its own script, with the advent of social media, code-switching has permeated into the Tamil language across informal contexts like forums and messaging outlets \cite{chakravarthi-etal-2019-wordnet,chakravarthi-etal-2018-improving, ghanghor-etal-2021-iiitk, ghanghor-etal-2021-iiitk-lt,yasaswini-etal-2021-iiitt}. As a result, code-switched content is part and parcel of offensive conversations in social media. 

Despite many recent NLP advancements, handling code-mixed offensive content is still a challenge in Dravidian Languages \cite{Sitaram2019ASO} including Tamil owing to limitations in data and tools. However, recently the research of offensive code-mixed texts in Dravidian languages has seen traction \cite{chakravarthi2021dataset,chakravarthi-etal-2020-corpus,priyadharshini2020named,chakravarthi-2020-hopeedi}. Yet, very few of these focus on identifying the spans that make a comment offensive \cite{ravikiran-annamalai-2021-dosa}. But accentuating such spans can help content moderators and semi-automated tools which prefer attribution instead of just a system-generated unexplained score per comment. Accordingly, in this shared task, we provided code-mixed social media text for the Tamil language with offensive spans inviting participants to develop and submit systems under two different settings. Our CodaLab website\footnote{\url{https://competitions.codalab.org/competitions/36395}} will remain open to foster further research in this area.

\begin{figure*}[]
    \centering
    \scalebox{0.9}{
    \includegraphics{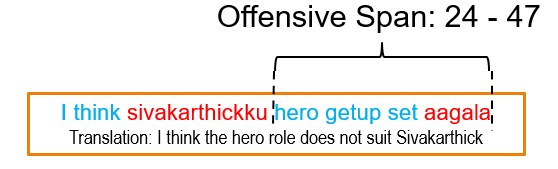}}
    \caption{Example Offensive Span Identification from Code-Mixed Tamil-English Text}
    \label{fig:x}
\end{figure*}

\begin{figure*}[]
    \centering
    \scalebox{0.8}{
    \includegraphics{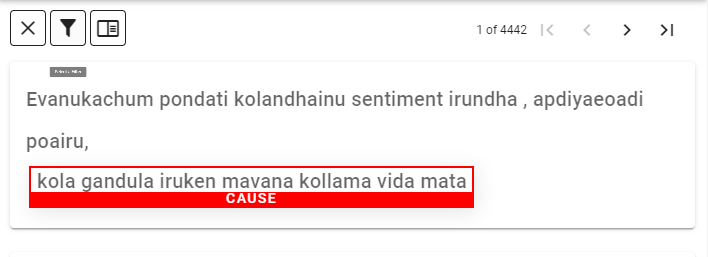}}
    \caption{Annotation of offensive spans using Doccano.}
    \label{fig:1}
\end{figure*}

\section{Related Work}

\subsection{Offensive Span Identification}

Much of the literature related to offensive span identification find their roots in SemEval Offensive Span identification shared task focusing on English Language \cite{pavlopoulos_androutsopoulos_sorensen_laugier}, with development of more than 36 different systems using a variety of approaches. Notable among these include work by \citet{zhu-etal-2021-hitsz} that uses token labeling using one or more language models with a combination of Conditional Random Fields (CRF). These approaches often rely on BIO encoding of the text corresponding to offensive spans. Alternatively, some systems employ post-processing on these token level labels, including re-ranking and stacked ensembling for predictions \cite{nguyen-etal-2021-nlp}. Then, there are exciting works of \citet{rusert-2021-nlp,plucinski-klimczak-2021-ghost} that exploit rationale extraction mechanism with pre-trained classifiers on external offensive classification datasets to produce toxic spans as explanations of the decisions of the classifiers. Lexicon-based baseline models, which uses look-up operations for offensive words \cite{burtenshaw-kestemont-2021-uantwerp} and run statistical analysis \cite{palomino-etal-2021-goldenwind} are also widely explored. Finally, there are a few approaches that employ custom loss functions tailored explicitly for false spans. For code-mixed Tamil-English to date, there is only preliminary work by \citet{ravikiran-annamalai-2021-dosa} that uses token level labeling.

\section{Task Description}
Our task of offensive span identification required participants to identify offensive spans i.e, character offsets that were responsible for the offensive of the comments, when identifying such spans was possible. To this end, we created two subtasks each of which are as described. Example of offensive span is shown in Figure \ref{fig:x}

\subsection{Subtask 1: Supervised Offensive Span Identification}
Given comments and annotated offensive spans for training, here the systems were asked to identify the offensive spans in each of the comments in test data. This task could be approached as supervised sequence labeling, training on the provided posts with gold offensive spans. It could also be treated as rationale extraction using classifiers trained on other datasets of posts manually annotated for offensiveness classification, without any span annotations.

\subsection{Subtask 2: Semi-supervised Offensive Span Identification}
All the participants of subtask 1 were also encouraged to submit a system to subtask 2 using semi-supervised approaches. Here in addition to training data of subtask 1, more unannotated data was provided. Participants were asked to develop systems using both of these datasets together. To this end, the unannotated data was allowed to be used in anyway as necessary to aid in overall model development including creating semi-supervised annotations, ranking based on similarity etc.

\section{Dataset}

For this shared task, we build upon dataset from earlier work of  \citet{ravikiran-annamalai-2021-dosa}, which originally released 4786 code-mixed Tamil-English comments with 6202 offensive spans. We released this dataset to the participants during training phase for model development. Meanwhile for testing we extended this dataset with new additional annotated comments. To this end, we use dataset of  \citet{DBLP:journals/corr/abs-2106-09460} that consist of 10K+ offensive comments. From this we filter out comments that were already part of train set resulting 4442 comments suitable for annotation. Out of this we created (a) 3742 comments were used for creating the test data and (b) 700 comments were used for training phase of subtask 2. 

\begin{table}[!htb]
\centering
\scalebox{0.7}{
\begin{tabular}{|c|c|c|}
\hline
\textbf{Split} & Train & Test \\ \hline
\textbf{Number of Sentences} & 4786 & 876 \\ \hline
\textbf{Number of unique tokens} & 22096 & 5362 \\ \hline
\textbf{Number of annotated spans} & 6202 & 1025 \\ \hline
\textbf{Average size of spans (\# of   characters)} & 21 & 21 \\ \hline
\textbf{Min size of spans (\# of   characters)} & 4 & 3 \\ \hline
\textbf{Max size of spans (\# of   characters)} & 82 & 85 \\ \hline
\textbf{Number of unique tokens in spans} & 10737 & 1006 \\ \hline
\end{tabular}}
\caption{Dataset Statistics used in this shared task}
\label{tab:2}
\end{table}

In line with earlier works \cite{ravikiran-annamalai-2021-dosa} for the 3742 comments we create span level annotations where at least two annotators annotated every comment. Additionally, we also employ similar guidelines for annotation, anonymity maintenance etc. Besides, no annotator data was collected other than their educational background and their expertise in the Tamil language. 

Additionally, all the annotators were informed in prior about the inherent profanity of the content along with an option to withdraw from the annotation process if necessary. For annotation, we  use doccano \cite{doccano}which was locally hosted by each annotator. Within doccano, all the annotators were explicitly asked to create a single label called \textbf{CAUSE} with label id of 1, thus maintaining consistency of annotation labels. (See Figure \ref{fig:1}). 

To ensure quality each annotation was verified by one or more annotation verifier, prior to merging and creating gold standard test set.  The overall dataset statistics is given in the Table \ref{tab:2}. Compared to train set, we can see that the test set consists of significantly lesser number of samples, this is because many of the comments were either small or were hard to clearly identify the offensive spans. Overall for the 876 comments we obtained Cohen's Kappa inter-annotator agreement of 0.61 inline with \citet{ravikiran-annamalai-2021-dosa}.

\section{Competition Phases}

\subsection{Training Phase}
In the training phase, the train split with 4786 comments, and their annotated spans were released for model development. Participants were given training data and offensive spans. No validation set was released; rather, participants were emphasized on cross-validation by creating their splits for preliminary evaluations or hyperparameter tuning. In total, 30 participants registered for the
task and downloaded the dataset.

\subsection{Testing Phase}
Test set comments without any span annotation were released in the testing phase. Each participating team was asked to submit their generated span predictions for evaluation. Predictions are submitted via Google form, which was used to evaluate the systems. Though CodaLab supports evaluation inherently, we used google form due to its simplicity. Finally, we assessed the submitted spans of the test set and were scored using character-based F1 (See section \ref{evalm}).

\section{System Descriptions}\label{desc}
Overall we received only a total of 4 submissions (2 main + 2 additional) from two teams out of 30 registered participants. All these were only for subtask 1. No submissions were made for subtask 2. Each of their respective systems are as described.

\subsection{The NITK-IT\_NLP Submission }\label{nitk_sub}

The best performing system from NITK-IT\_NLP \cite{hariharan-acl} experimented with rationale extraction by training offensive language classifiers and employing model-agnostic rationale extraction mechanisms to produce toxic spans as explanations of the decisions of the classifier. Specifically NITK-IT\_NLP used MuRIL \cite{DBLP:journals/corr/abs-2103-10730} classifier and coupled with LIME \cite{DBLP:conf/naacl/Ribeiro0G16} and used the explanation scores to select words suitable for offensive spans.

\subsection{The DLRG submission}\label{dlrgsub}
The DLRG team \cite{vit-acl} formulated the problem as a combination of token labeling and span extraction. Specifically, the team created word-level BIO tags i.e., words were labelled as B (beginning word of a offensive span), I (inside word of a
offensive span), or O (outside of any offensive span). Following which word level embeddings are created using GloVe \cite{pennington-etal-2014-glove} and BiLSTM-CRF \cite{panchendrarajan-amaresan-2018-bidirectional} model is trained.

\subsection{Additional Submission}
After testing phase, we also requested each team to submit additional runs if they have variants of approaches. Accordingly we received two additional submissions from NITK-IT\_NLP where they replaced MuRIL from their initial submission with (i) Multilingual-BERT \cite{Devlin2019BERTPO} and (ii) ELECTRA \cite{DBLP:conf/iclr/ClarkLLM20} respectively without any other changes. More details in section \ref{evalm}.

\section{Evaluation}

This section focuses on the evaluation framework
of the task. First, the official measure that was
used to evaluate the participating systems is described. Then, we discuss baseline models that were selected as benchmarks for comparison reasons. Finally, the results are presented.

\subsection{Evaluation Measure}
In line with work of \citet{pavlopoulos_androutsopoulos_sorensen_laugier} each system was evaluated F1 score computed on character offset. For each system, we computed the F1 score per comments, between the predicted and the ground truth character offsets. Following this we calculated macro-average score over all the 876 test comments. If in case both ground truth and predicted character offsets were empty we assigned a F1 of 1 other wise 0 and vice versa.

\subsection{Benchmark}\label{evalm}
To establish fair comparison we first created two baseline benchmark systems which are as described.

\begin{itemize}
    \item \texttt{BENCHMARK 1} is a random baseline model which randomly labels 50\% of characters in comments to belong to be offensive. To this end, we run this benchmark 10 times and average results are presented in Table \ref{tab:x}.
    \item \texttt{BENCHMARK 2} is a lexicon based system, which first extracted all the offensive words from the train set and during inference these words were searched in comments from testset and corresponding spans were extracted.
     \item \texttt{BENCHMARK 3} is RoBERTA \cite{DBLP:journals/corr/abs-1907-11692,ravikiran-annamalai-2021-dosa} model trained using token labeling approach with BIO encoded texts corresponding to annotated spans. 
\end{itemize}

\begin{table}[!htb]
\centering
\caption{Official rank and F1 score (\%) of the 2 participating teams that submitted systems. The baselines benchmarks are shown in red.}
\label{tab:x}
\scalebox{0.8}{
\begin{tabular}{|c|c|c|}
\hline
\textbf{RANK} & \textbf{TEAM} & \textbf{F1 (\%)} \\ \hline
1 & NITK-IT\_NLP & 44.89 \\ \hline
{\color[HTML]{000000} \texttt{BASELINE}} & {\color[HTML]{000000} \texttt{BENCHMARK 1}} & {\color[HTML]{000000} 39.75} \\ \hline
{\color[HTML]{000000} \texttt{BASELINE} }& {\color[HTML]{000000} \texttt{BENCHMARK 2}} & {\color[HTML]{000000} 37.84} \\ \hline
{\color[HTML]{000000} \texttt{BASELINE} }& {\color[HTML]{000000} \texttt{BENCHMARK 3}} & {\color[HTML]{000000} 38.61} \\ \hline
2 & DLRG & 17.28 \\ \hline
\end{tabular}}
\end{table}

Table \ref{tab:x} shows the scores and ranks of two teams that made their submission. NITK-IT\_NLP (Section \ref{nitk_sub}) was ranked first, followed by DLRG
(Section \ref{dlrgsub}) that scored 27\% lower was ranked second. The median score was 31.08\%, which is far below the top ranked team and the benchmark baseline models. Meanwhile the additional submission post testing phase are excluded from ranked table. Instead they are presented separately in Table \ref{tab:z}.

\texttt{BENCHMARK 1}  achieves a considerably high score and, hence, is very highly ranked with character F1 of 39.83\%. Combination of MuRIL with LIME interpretability by model NITK-IT\_NLP is ahead of \texttt{BENCHMARK 1} by 11\%, indicating the language models ability to effectively rationalize and identify the spans. This is inline the results of \citet{rusert-2021-nlp} which show higher results than random baseline. Meanwhile \texttt{BENCHMARK 2} and \texttt{BENCHMARK 3}, also shows  F1 of 37.84\% and 38.61\% which again NITK-IT\_NLP model tend to beat significantly. On contrary we could see that DLRG model to show least results of 17.28\% lesser than akk the baselines as well as the top performing system. The lexicon-based \texttt{BENCHMARK 2} and RoBERTA based \texttt{BENCHMARK 3} too score very high. Especially as it overcomes, the submission of DLRG. This may be attributed to dataset domain itself. Especially, since much of the dataset was collected from Youtube comments section of Movie Trailers, often we see usages of same word or similar words. Such behavior is well established across social media forums including Youtube \cite{DBLP:journals/corr/abs-2109-01127}, which begs to ask if indeed the dataset construction needs to be revisited, which forms one potential exploration for immediate future.

\section{Analysis and Discussion}

Overall we were happy to see the degree of involvement in this shared task with multiple participants registering, requesting access to datasets and potential baseline codes for the shared task. Though only two teams submitted the systems, the resulting diversity of approaches to this problem is fairly encouraging. However we include some of our observations below, from our evaluation and what we have learned from the results.

\begin{table}[!htb]
\centering
\caption{Results of additional runs submitted by NITK-IT\_NLP.}
\label{tab:z}
\scalebox{0.8}{
\begin{tabular}{|c|c|}
\hline
\textbf{Method} & \multicolumn{1}{c|}{\textbf{F1 (\%)}} \\ \hline
ELECTRA + LIME & 37.33 \\ \hline
M-BERT + LIME & 33.95 \\ \hline
\end{tabular}}
\end{table}

\subsection{Participation Characteristics}

The authors reached out to teams that initially registered but failed to create any systems and the vast majority were undergraduate students who were new into the concept of shared task and were time-limited due to semester exams. The fact that students participated in the task is promising
and we plan to consider more ways to introduce Shared tasks on Low-Resource Dravidian Languages in classrooms. To this end, the we used social media and other medium to spread the word around universities.

On the other hand, 60\% of the participants did not download dataset after registering and instead chose to participate in other shared tasks, which is problematic and should be addressed. To this end, correspondence with such teams revealed potential favoritism towards classification based problems that are common in undergraduate studies. Moreover we also received multiple queries on the concept of offensive span itself during the training phase, which is a indicates potential need of improving the overall task structure with potential early release of data and task details. Yet, upon extending the number of submissions NITK-IT\_NLP submitted additional runs (See Table \ref{tab:z}). Additionally both the teams also submitted source codes \footnote{\url{https://drive.google.com/drive/folders/1T3kl8mljPt8oXcKVn7OQqaU3d55za2zZ?usp=sharing}} for their respective models encouraging further development of systems.

\subsection{General remarks on the approaches}

Though neither of teams that made final submissions created any simple baselines, we could see that all the submissions of NITK-IT\_NLP use well established approaches in recent NLP focusing on pretrained language models. Meanwhile DLRG used well-grounded  Non-Transformer based approach. Yet neither of teams used any ensembles, data augmentation strategies or modifications to loss functions that are seen for the task of span identification in the past across shared tasks.  

\subsection{Error Analysis}

Table \ref{tab:x} shows maximum result of $0.4489$ with DLRG failing significantly compared to random baseline. To this end, we wonder if potentially these approaches have any weaknesses or strengths. To understand this, first we study the character F1 results across sentences of different lengths. Specifically we analysis results of (a) comments with less than 30 characters (F1@30) (b) comments with 30-50 characters (F1@50) (c)  comments with more than 50 characters (F1@>50). The results so obtained are as shown in Table \ref{tab:y}.

\begin{table}[]
\centering
\caption{Results of submitted systems across comments of different lengths.}
\label{tab:y}
\scalebox{0.7}{
\begin{tabular}{|c|c|c|c|}
\hline
 & \textbf{F1@30 (\%)} & \textbf{F1@50 (\%)} & \textbf{F1@\textgreater{}50 (\%)} \\ \hline
\textbf{NITK-IT\_NLP} & 42.39 & 37.05 & 26.42 \\ \hline
\textbf{DLRG} & 39.62 & 23.47 & 14.05 \\ \hline
\end{tabular}}
\end{table}

Firstly we can see though NITK-IT\_NLP shows high results overall for cases of comments with larger lengths the model fails significantly. Specifically, comparing results with ground truth showed that use of LIME often restricts the overall word so selected as the rationale for offensiveness in turn reducing number of character offsets predicted as spans. This is because  with larger texts the net score distribution weakens and span extraction is largely off leading to significant drop in results. Meanwhile for DLRG the results are more mixed, especially we can see that for comments with less than 30 characters the model shows improvement in F1. Analysis of results reveal that token labeling is highly accurate, which drops significantly with large size sentences. This may be attributed to non-local interactions between the words that may not be captured by the Bi-LSTM CRF model. Further more much of these sentences often contained only cuss words or clearly abusive words that are easily identifiable and often present in the train set. Also we found few bugs in the training code so used, which was already informed to the authors.

Besides error analysis also showed some implicit challenges in the proposed shared task. First the strong dependency of offensiveness on context makes it particularly difficult to solve as evident from NITK-IT\_NLP which used language models. Second, offensiveness often is expressed as sarcasm or even is very subtle. In such cases we often see the offensiveness results to depend only the words bearing the most negative sentiment, meanwhile the ground truth spans annotated are larger thus showing high errors. Finally, many times the nature of offensiveness itself becomes debatable without clear context. Often these are the cases where we find the developed approaches to fail significantly.

\section{Conclusion}

Overall this shared task on offensive span identification we introduced a new dataset for code-mixed Tamil-English language with total of 5652 social media comments annotated for offensive spans. The task though has large participants, eventually had only two teams that submitted their systems. In this paper we described their approaches and discussed their results.  Surprisingly rationale extraction based approach involving combination MuRIL and LIME performed significantly well. Meanwhile Bi-LSTM CRF model was found showing sensitivity towards shorter sentences, though it performed significantly worse than the random baseline. Also extracting offensive spans for long sentences were found to be difficult especially as they are context dependent. To this end, we release the baseline models and datasets to foster further research. Meanwhile in the future we plan to re-do the task of offensive span identification where we could require the participants to identify offensive spans and simultaneously classify different types of offensiveness.  

\section*{Acknowledgements}
We thank our anonymous reviewers for their valuable feedback. Any opinions, findings, and conclusion or recommendations expressed in this material are those of the authors only and does not reflect the view of their employing organization or graduate schools. The shared task was result of series projects done during CS7646-ML4T (Fall 2020), CS6460-Edtech Foundations (Spring 2020) and  CS7643-Deep learning (Spring 2022) at Georgia Institute of Technology (OMSCS Program). Bharathi Raja Chakravarthi were supported in part by a research grant from Science Foundation Ireland (SFI) under Grant Number SFI/12/RC/2289$\_$P2 (Insight$\_$2), co-funded by the European Regional Development Fund and Irish Research Council grant IRCLA/2017/129 (CARDAMOM-Comparative Deep Models of Language for Minority and Historical Languages).

\bibliography{anthology,custom}
\bibliographystyle{acl_natbib}

\end{document}